\pdfoutput=1

\documentclass[11pt]{article}
\usepackage{authblk}

\usepackage{pdfpages}
\usepackage{subfig}
\usepackage{makecell}
\usepackage[]{emnlp2021}

\usepackage{times}
\usepackage{latexsym}

\usepackage[T1]{fontenc}
\usepackage{tikz}

\usepackage[utf8]{inputenc}

\usepackage{microtype}

\usepackage{microtype}
\usepackage{graphicx}
\usepackage{booktabs} 
\usepackage{times}
\usepackage{latexsym}
\usepackage{multirow}
\usepackage{array}
\newcolumntype{L}{>{\arraybackslash}m{12cm}}
\usepackage{amssymb}
\usepackage{physics}

\makeatletter
\newcommand\blfootnote[1]{%
  \begingroup
  \renewcommand{\@makefntext}[1]{\noindent\makebox[1.8em][r]#1}
  \renewcommand\thefootnote{}\footnote{#1}%
  \addtocounter{footnote}{-1}%
  \endgroup
}
\makeatother

\title{FrugalScore: Learning Cheaper, Lighter and Faster Evaluation Metrics for Automatic Text Generation}

\author[1]{Moussa Kamal Eddine}
\author[2*]{Guokan Shang}
\author[1*]{Antoine J.-P. Tixier}
\author[1]{Michalis Vazirgiannis}
\affil[1]{\'Ecole Polytechnique, $^\mathrm{2}$Linagora}

\begin{document}
\maketitle
\begin{abstract}
\blfootnote{*Equal contribution}
Fast and reliable evaluation metrics are key to R\&D progress.
While traditional natural language generation metrics are fast, they are not very reliable.
Conversely, new metrics based on large pretrained language models are much more reliable, but require significant computational resources.
In this paper, we propose FrugalScore, an approach to learn a fixed, low cost version of any expensive NLG metric, while retaining most of its original performance.
Experiments with BERTScore and MoverScore on summarization and translation show that FrugalScore is on par with the original metrics (and sometimes better), while having several orders of magnitude less parameters and running several times faster.
On average over all learned metrics, tasks, and variants, FrugalScore retains 96.8\% of the performance, runs 24 times faster, and has 35 times less parameters than the original metrics. 
We make our trained metrics publicly available\footnote{https://github.com/moussaKam/FrugalScore}, to benefit the entire NLP community and in particular researchers and practitioners with limited resources.

\end{abstract}

\section{Introduction}
Automatic evaluation metrics are the only way to monitor the training of, evaluate, and compare across models in a systematic, large-scale way, and are thus a critical component of the research and development ecosystem in machine learning.
To get adopted in practice, evaluation metrics need to be both reliable and affordable, i.e., fast and easy to compute.

While some metrics meet these criteria, such as precision and recall in information retrieval, root mean square error in regression, etc., finding suitable metrics is still an open problem in the field of Natural Language Generation (NLG) \cite{novikova2017we}.

Indeed, historical $n$-gram matching metrics such as ROUGE \citep{lin2004rouge} for summarization, BLEU \citep{papineni-etal-2002-bleu} and METEOR \citep{banerjee-lavie-2005-meteor} for translation, while affordable, are not very reliable, as they are based on surface-form matching only, i.e., lexical similarity, and have thus no sense of semantic similarity.
For instance, it makes little sense to use ROUGE for the evaluation of abstractive summarization systems (which are becoming the norm), or whenever the generated text paraphrases the original text. 

Following the advent of transfer learning in NLP, new NLG metrics based on large pretrained language models have recently been proposed, such as BERTScore \cite{zhang2019bertscore} and MoverScore \cite{zhao2019moverscore}.
By relying on contextual embeddings, these metrics capture semantics and are therefore much more reliable.
However, due to the sheer size of the underlying models, these metrics pose environmental issues \cite{strubell2019energy}, take time to compute, and require access to significant computational resources, so they are not accessible by everyone in the NLP community.

For example, we were not able to run some of the best variants of BERTScore\footnote{From BERTScore's authors: https://tinyurl.com/8cwyter2}, based on DeBERTa-Large and DeBERTa-XLarge \cite{he2020deberta} on a 12GB GPU.
Even when enough GPU memory is available, relying on such large models is still associated with extended runtimes, which can impede the progress of experiments when used once or more per epoch for validation and monitoring purposes. 

To address this problem, we propose in this paper FrugalScore, an approach to learn a lightweight version of BERTScore, MoverScore, and more generally any metric based on a large pretrained language model.
While our objectives are clearly the same as that of model compression, and distillation in particular, our method differs: we first sample sequence pairs, annotate these pairs with the metric to be learned, and finally train a miniature model on the resulting dataset.

Our contributions can be summarized as follows:

\noindent 1) Our compact models have several orders of magnitude less parameters than the original metrics and run several times faster, while retaining most of their original performance.
We even outperform the original metrics in some cases\footnote{Hence the name FrugalScore, as frugal engineering is defined as ``achieving more with fewer resources''.}.

\noindent 2) Our metrics are not only faster because of the much smaller amount of parameters, but also because they require only one forward pass and do not rely on any similarity function.

\noindent 3) Regardless of how expensive the original metric is, querying our trained metrics always has the same low, fixed cost.
This decoupling is a major advantage as the size of the pretrained language models has recently been growing tremendously (e.g., \citet{brown2020language}).

\section{Background}\label{sec:background}
Related work falls into two categories: unsupervised and supervised metrics.

\subsection{Unsupervised metrics}
To address the limitations of ROUGE and BLEU, variants based on static word embeddings \citep{mikolov2013efficient} were developed, e.g., ROUGE-WE \citep{ng2015better}, BLEU2VEC \cite{tattar2017bleu2vec}, and MEANT 2.0 \citep{lo-2017-meant}.
While using word vectors is a progress over strict $n$-gram matching, static embeddings are still very limited as they do not capture polysemy, i.e., the fact that words have different meanings in different contexts.

More recently, the focus has shifted to harnessing the power of the contextualized embeddings produced by large pretrained language models.
For instance, the Sentence Mover's Similarity \citep{clark2019sentence} represents sentences as the average of their ELMo word embeddings \citep{peters-etal-2018-deep} and measures the minimum cost of transforming one summary into the other, using a modified version of the Word Mover's Distance \citep{kusner2015word}.
\textsc{BERTr} \citep{mathur2019putting} computes approximate recall based on the pairwise cosine similarity between the BERT embeddings \citep{devlin2018bert} of the words in automatic and reference translations.
Mark-Evaluate \citep{mordido-meinel-2020-mark} is a family of metrics that consider contextualized word or sentence embeddings derived from BERT as population samples, to evaluate language generation with population estimation methods used in ecology.

Finally, the recently introduced BERTScore \cite{zhang2019bertscore} and MoverScore \cite{zhao2019moverscore} are general-purpose NLG evaluation metrics that are becoming widely used.
The main difference between BERTScore and MoverScore lies in the function used to compute the similarity between the representations of the two sequences $\mathbf{x}= \langle \mathbf{x_1},..., \mathbf{x_k} \rangle$ and  $\mathbf{y}= \langle \mathbf{y_1},..., \mathbf{y_l} \rangle$.
We experimented with these two metrics, so we provide more details about them in what follows.\\

\noindent\textbf{BERTScore} first computes the pairwise cosine similarity between the representations of the tokens in each sequence, and uses greedy matching to match each token to the most similar one in the other sequence.
Given two pre-normalized vector sequences $\mathbf{x}$ and $\mathbf{y}$, BERTScore computes:

\begin{equation}
R_{BERT}=\frac{1}{|\mathbf{x}|}\sum_{\mathbf{x_i} \in x}\max_{\mathbf{y}_j \in \mathbf{y}}\vb{\mathbf{x^T_i}\mathbf{y}_j }
\end{equation}
and:
\begin{equation}
P_{BERT}=\frac{1}{|\mathbf{y}|}\sum_{\mathbf{y_i} \in y}\max_{\mathbf{x}_j \in \mathbf{x}}\vb{\mathbf{y^T_i}\mathbf{x}_j}\end{equation}
The F1-score is classically obtained as: 
\begin{equation}
F_{BERT} = 2 \frac{P_{BERT} R_{BERT}}{P_{BERT} + R_{BERT}}
\end{equation}

\noindent\textbf{MoverScore} uses an $n$-gram generalization of the Word Mover's Distance (WMD) \cite{kusner2015word} as their (dis)similarity function.
More specifically, they solve for the optimal transportation flow matrix $F \in \mathbb{R}^{|\mathbf{x}| \times |\mathbf{y}|}$ between the two weighted sequences of $n$-grams:

\begin{equation}
WMD(\mathbf{x}, \mathbf{y}) = min_{F}\langle C,F \rangle
\end{equation}
$$s.t.\quad  F\mathbf{1} = f_{\mathbf{x}},~~ F^T\mathbf{1}=f_{\mathbf{y}}$$
Where $C$ is the transportation cost matrix ($C_{ij}$ is the Euclidean distance between $x_i$ and $y_j$) and $f_x \in \mathbb{R}^{|\mathbf{x}|}_+ $ and $f_y \in \mathbb{R}^{|\mathbf{y}|}_+$ are the $n$-gram weight vectors.\\

Note that by directly learning BERTScore's and MoverScore's full internal mapping (from sequence pairs to final scalar scores), FrugalScore internalizes their similarity functions.
This does not only provide a speedup at inference time, but also improves performance, as shown in section \ref{sec:results}.

\subsection{Supervised metrics}\label{sec:supmet}
Related to our work are also supervised metrics, which are directly trained on human evaluations.
ROSE \citep{conroy-dang-2008-mind} is a linear combination model of different variants of ROUGE using canonical correlation.
\textsc{Beer} \citep{stanojevic2014beer} is a learning-to-rank approach using word and character n-gram matching, and token ordering, as features to maximize correlation with human rankings of machine translation systems.
S$^3$ \citep{peyrard-etal-2017-learning} trains a regression model that takes the evaluation scores of several existing metrics and many hand-crafted features as input, and learns the best combination of them to approximate human summary judgments.
DPMFcomb \citep{yu-etal-2015-casict} and Blend \citep{ma-etal-2017-blend} are combined metrics incorporating a vast amount of lexical, syntactic and semantic based translation evaluation metrics using ranking and regression SVMs respectively.
RUSE \citep{shimanaka-etal-2018-ruse} evaluates machine translation with a neural regressor based on universal sentence embeddings (e.g., InferSent \citep{conneau-etal-2017-supervised}).
NUBIA \citep{kane-etal-2020-nubia} consists of three modules: a feature extractor based on RoBERTa \citep{liu2019roberta} and GPT-2 \citep{radford2019language} fine-tuned on language evaluation tasks, an aggregator trained to predict the quality of the hypothesis given the reference using the extracted features, and a calibrator mapping all predictions between 0 and 1.

\noindent\textbf{Differences}. Like the aforementioned efforts, FrugalScore is a learned metric.
However, it does not rely on any intermediate or handcrafted features, and, most importantly, it does not require training on human annotations.
Supervision in FrugalScore is conducted on a synthetic dataset, as a trick to expose and learn the internal mapping of the unsupervised metrics to be learned.
Last but not least, unlike all aforementioned methods, compression is central to FrugalScore, which is based on miniature versions of the models used by the original metrics.


\subsection{Differences with distillation}
Knowledge distillation \citep{hinton2015distilling} is the process of transferring knowledge from a large teacher model to a smaller student model to accomplish model compression \citep{bucilua2006model}.
It was originally proposed in the domain of computer vision and speech recognition, then successfully adapted to NLP \citep{sanh2019distilbert}.
While FrugalScore, like distillation, focuses on model compression, there is one major difference.
Distillation was designed for multi-class classification settings, and relies on a cross-entropy loss over the softened probability distributions of the teacher and student.
In our case, we deal with a regression setting, and use the mean squared error objective. 

\subsection{Differences with BLEURT}
A work closely related to ours is BLEURT \cite{sellam2020bleurt}. 
However, there are a number of significant differences with our approach.
First, BLEURT continues the pretraining of an already pretrained BERT-based model on a synthetic dataset in a self-supervised way, whereas FrugalScore is directly trained to learn the scores of the metric of interest, in a supervised fashion.

Also, BLEURT's synthetic dataset is made by perturbing Wikipedia sentences with mask-filling, backtranslation, and word dropping, whereas we use other data sources than Wikipedia such as summarization and translation datasets, and only NLG models to induce perturbations.

When creating its synthetic dataset, BLEURT automatically annotates the (original, perturbed) sequence pairs with numerical and categorical ``signals'': BLEU, ROUGE, BERTscore, backtranslation likelihood, textual entailment (probability of three labels: entail, contradict, and neutral, given by BERT fine-tuned on MNLI), and backtranslation flag.
On the other hand, FrugalScore simply and directly annotates the sequence pairs with the metric to be learned.

After pretraining, BLEURT is fine-tuned on human judgments, in a way similar to the supervised metrics described in subsection \ref{sec:supmet}.
BLEURT does not learn to generate a scalar until that final fine-tuning phase, so it cannot be used as a metric before that.
Conversely, FrugalScore is trained from the start to be a metric, and the fine-tuning phase is optional.

Also, BLEURT was designed for the evaluation of translation.
The authors only test whether it can be applied to a different task by experimenting on the WebNLG (data-to-text) dataset \citep{gardent-etal-2017-webnlg}.
Conversely, we focus on learning general text similarity metrics (e.g., BERTscore and MoverScore), so FrugalScore is task-agnostic by design.

Finally, and above all, the objective of FrugalScore is model compression, whereas that of BLEURT is metric learning.


\section{Our Approach}
Developing FrugalScore requires three phases, one of which is optional.

\noindent \textbf{Phase 1}.
We create a synthetic dataset (see subsection \ref{sub:synth}) by sampling pairs of more or less related sequences and annotating them with the expensive metrics to be learned.
This is a one-time operation that does not need to be repeated regardless of the model used in Phase 2.

\noindent \textbf{Phase 2}. 
We continue the pretraining (subsection \ref{subsec:pretraining}) of a miniature pretrained language model on the synthetic dataset built by Phase 1.
Here, the miniature model learns the internal mapping of the expensive metric, including any similarity function applied to the representations.
Note that a different miniature is trained for each metric to be learned (we leave learning metric combinations as future work).

The miniature can then be used in inference mode to generate scores for any never-seen pair of sequences.

\noindent \textbf{Phase 3} (optional).
We fine-tune the miniature on human annotations, which, as shown in section \ref{section:supervised}, can boost performance.\\


\subsection{Synthetic Dataset}\label{sub:synth}
The objective here was to generate pairs of sequences  mimicking the (reference, candidate) pairs found in NLG datasets, which are usually semantically related and in many cases paraphrasing one another. 
We sampled our sequences from a variety of data sources, listed next.

\noindent \textbf{Summarization}.
For each document in the well-known CNN/DailyMail dataset \cite{nallapati2016abstractive}, our goal was to generate several summaries differing in terms of structure and quality.
To this purpose, we used different pretrained seq2seq summarization models: BART-base and BART-large \cite{lewis2019bart}, mBART \cite{liu2020multilingual}, and BARThez \cite{eddine2020barthez}.
BART is a seq2seq autoencoder with a Transformer architecture.


The four models were fine-tuned for one epoch on 50k examples randomly sampled from the training set of CNN/DM, and were used to generate summaries for the whole training set of 287,112 documents, using greedy decoding.

Note that we kept the 50K documents used for fine-tuning in the final generation pool, in order to create quality differences among summaries.
Indeed, models are expected to better summarize the documents used for training than never-seen documents.

We also used the human reference summaries, so that in the end, each document was associated with 5 summaries, resulting in 10 pairs of summaries per document.

\noindent \textbf{Backtranslation}.
We also generated paraphrases with backtranslation, by sampling sentences from the OpenSubtitle English monolingual corpus \cite{lison2016opensubtitles2016}, and translating them to French, Arabic and German with OPUS-MT \cite{TiedemannThottingal:EAMT2020},
before translating them back to English.
We used OPUS-MT because of its ready-to-use checkpoints available for many language pairs.
We ended up with 4 variations for each sentence (including the original one), resulting in 6 paraphrase pairs per sentence.

\noindent \textbf{Denoising}.
To avoid bias towards summarization and translation, we also generated pairs of related sequences such that the first element in the pair was a Wikipedia segment and the second element was a BART-denoised version of it \cite{lewis2019bart}.

More precisely, we sampled 2M segments from Wikipedia such that the number of unigrams in these segments was uniformly distributed between 1 and 200.
Our assumption was that enforcing variations in sequence length would help the learned metric to generalize.

We then applied BART's \textit{text infilling} and \textit{sentence permutation} perturbation strategies to each segment.
That is, multiple text spans were sampled and replaced with a \texttt{[MASK]} special token.
The lengths of the spans were sampled from a Poisson distribution ($\lambda=3$).
$50\%$ of the tokens within the input segment were masked and $20\%$ of the masked text was replaced with random tokens (creating pathological examples to increase the robustness of the learned metric).
The sentences in the input segment were then shuffled.

We finally used a BART-Base checkpoint\footnote{https://dl.fbaipublicfiles.com/fairseq/models/bart.base.tar.gz} from the Fairseq library \cite{ott2019fairseq} to try to reconstruct the perturbed versions of the original sequences, hence creating variants of them.


\noindent \textbf{Annotating pairs}.
We sampled 4.5M sequence pairs uniformly from each aforelisted source.
These pairs were then annotated with the metrics to be learned.
Note that this is a one-time operation that does not need to be repeated regardless of which models are trained downstream.

In this work, we experimented with two recent expensive NLG metrics that rely on large pretrained language models, BERTScore \cite{zhang2019bertscore} and MoverScore  \cite{zhao2019moverscore}, presented in section \ref{sec:background}.
However, it is important to note that our method can be used with any other NLG metric.

Note that for BERTScore, we used the F-1 score $F_{BERT}$, as recommended by the authors \cite{zhang2019bertscore}.
For MoverScore, still following the authors \cite{zhao2019moverscore}, we used the variant operating on unigrams and the IDF to compute the vectors of weights.

\begin{table*}[ht]
\small
\def\arraystretch{1.5}
\begin{center}
\begin{tabular}{  |c|c|c|c|c|c|c|c|c|c|c|c|c|} 
\hline
 & Metric & Model & \makecell{Scores \\ (TAC)} & \makecell{Runtime \\ (TAC)} & \makecell{Scores \\ (WMT)} & \makecell{Runtime \\ (WMT)} & Params \\
\hline
a & BERTScore & BERT-Tiny  & 55.4/47.5 & 1m 27s & 37.6 & 1m 22s & 4.4M \\
b & BERTScore & BERT-Small & 61.6/51.5 & 2m 20s & 39.1 & 1m 42s & 29.1M\\
c & BERTScore & BERT-Medium & 62.7/52.4 & 2m 28s & 39.8 & 2m 04s & 41.7M \\
\hline
d & BERTScore & BERT-Base & 64.7/54.7 & 3m 28s & 41.9 & 2m 09s & 110M \\
e & BERTScore & RoBERTa-Large & 64.2/55.4 & 5m 17s & 43.2 & 3m 03s & 355M \\
f & BERTScore & DeBERTa-XLarge & 64.5/\textbf{56.0} & 6m 20s & \textbf{44.5} & 3m 49s & 900M \\
g & MoverScore & BERT-Base & \textbf{66.5}/55.4 & 301m 29s& 44.0 & 64m 32s & 110M \\
\hline \hline
i & FrugalScore$_d$ & BERT-Tiny & 64.9/53.5 & 1m 28s & 38.4 & 1m 18s & 4.4M  \\
ii & FrugalScore$_d$ & BERT-Small & 64.7/53.7 & 2m 29s & 41.3 & 1m 35s & 29.1M \\
iii & FrugalScore$_d$ & BERT-Medium & 64.8/54.2 & 3m 41s & 41.9 & 1m 55s & 41.7M \\
\hline
iv & FrugalScore$_e$ & BERT-Tiny & 60.0/50.1 & 1m 28s & 37.5 & 1m 18s & 4.4M \\
v & FrugalScore$_e$ & BERT-Small &  64.1/53.8 & 2m 29s & 40.5 & 1m 35s & 29.1M \\
vi & FrugalScore$_e$ & BERT-Medium & 63.9/52.1 & 3m 41s & 41.7 & 1m 55s & 41.7M \\
\hline
vii & FrugalScore$_f$ & BERT-Tiny & 61.7/51.0 & 1m 28s & 38.0 & 1m 18s & 4.4M \\
viii & FrugalScore$_f$ & BERT-Small & 66.0/54.9 & 2m 29s & 41.5 & 1m 35s & 29.1M \\
ix & FrugalScore$_f$ & BERT-Medium & 65.5/54.9 & 3m 41s & 43.0 & 1m 55s & 41.7M \\
\hline
x & FrugalScore$_g$ & BERT-Tiny & \textbf{67.3}/\textbf{55.1} & 1m 28s & 39.8 & 1m 18s & 4.4M \\
xi & FrugalScore$_g$ & BERT-Small & 65.9/54.7 & 2m 29s & 42.8 & 1m 35s & 29.1M \\
xii & FrugalScore$_g$ & BERT-Medium & 66.2/\textbf{55.1} & 3m 41s & \textbf{43.6} & 1m 55s & 41.7M \\
\hline
\end{tabular}
\end{center}
\caption{\label{tab:evaluation} Scores are summary-level (TAC) and segment-level (WMT) Pearson correlations averaged over 2008 to 2011 for TAC (pyramid score/responsiveness) and over all source languages for WMT-2019.
Runtimes include preprocessing.
Subscripts refer to row labels and indicate which metric-model combination was used to annotate pairs (e.g., for FrugalScore$_d$, it is row $d$, i.e., BERTScore-BERT-Base). 
}
\end{table*}

\subsection{Metric Learning}\label{subsec:pretraining}
We continue the pretraining of three BERT miniatures\footnote{https://huggingface.co/google} on our synthetic dataset: BERT-Tiny ($L=2$, $H=128$), BERT-Small ($L=4$, $H=512$) and BERT-Medium ($L=8$, $H=512$), where $L$ is the number of layers and $H$ is the dimension of the embedding space.
These models have respectively 25 times, 3.78 times, and 2.64 times less parameters than BERT-base.
The concept of BERT miniatures was introduced by \citet{turc2019well} to test whether pretraining small models from scratch was competitive to distilling very large models. 
The miniature models have already been pretrained on masked language model and next sentence prediction objectives.

We continue pretraining using the standard method introduced by \citet{devlin2018bert}.
We concatenate the two sequences $x = \langle x_1,...,x_k \rangle$ and $y = \langle y_1,...,y_l \rangle$ in a given pair, separating them with a special \texttt{[SEP]} token.
A special \texttt{[CLS]} token is also added at the beginning of the resulting sequence.
The sequence of contextualised embeddings $\langle \mathbf{z_{[CLS]}}, \mathbf{x_1}, ... \mathbf{x_k}, \mathbf{z_{[SEP]}}, \mathbf{y_1}, ... , \mathbf{y_l} \rangle$ is then obtained.
We finally add a fully connected layer on top, that linearly projects the $\mathbf{z_{[CLS]}}$ vector to a scalar $s$.

The model is trained to minimize the mean square error (MSE) loss between the learned metric $s_i$ and the metric to be learned $\hat{s_i}$ (i.e., the annotation of the pair):

\begin{equation}
l = \frac{1}{N}\sum_{n=1}^{N}||s_i - \hat{s_i}||^2
\end{equation}
When pretraining is over, the models can be further fine-tuned on smaller human-annotated datasets as shown in section \ref{section:supervised}, or directly used to generate scores for unseen examples as shown in section \ref{experiments}. \\

\noindent \textbf{Setup}.
We use a batch size of 32 and the Adam optimizer \cite{kingma2014adam} with a learning rate of $3\times10^{-5}$,
linear decay, and a warm-up for 6\% of the total training steps, and we train the model for three epochs.
We conducted the pretraining on a single TITAN RTX GPU (24GB).
It took 10, 24 and 33 hours, respectively for the tiny, small, and medium miniatures.
We rely on the Transformers library \cite{wolf2019huggingface} for all pretraining and fine-tuning experiments.

\section{Experiments}\label{experiments}
In this section, FrugalScore is used in inference mode to generate scores directly after pretraining, i.e., no fine-tuning is performed (see section \ref{section:supervised} for fine-tuning results).

We evaluate on two text generation tasks: summarization and translation.
We use evaluation datasets containing (reference, candidate) sequence pairs annotated with human scores assessing the quality of the candidates given the references.
We measure the effectiveness of FrugalScore by measuring the Pearson correlation of its scores with the human judgments and comparing it to that of the original metrics.
We also take the number of parameters and the runtime into account.

\noindent \textbf{Text Summarization}.
We use 4 different multi-document summarization datasets from the Text Analysis Conference (TAC)\footnote{https://tac.nist.gov/}: TAC-2008, TAC-2009, TAC-2010 and TAC-2011.

These datasets respectively contain 48, 44, 46 and 44 clusters of documents and 58, 55, 43 and 51 systems are used to generate summaries.
Each cluster forms a topic to be summarized and has 4 reference summaries.
There are approximately 10k pairs in each dataset.
Each pair is annotated with two human judgment scores: the \textit{Pyramid Score} \cite{harnly2005automation} and the \textit{Responsiveness} \cite{dang2008overview}. The former measures the proportion of important semantic units (SCUs) in the reference summaries captured by the system summary, while the latter reflects the content coverage and the readability of each summary. 

\noindent \textbf{Machine Translation}.
Our evaluation corpus is from the WMT-2019\footnote{http://www.statmt.org/wmt19/} shared task \cite{li2019findings}. 
We consider all the to-English pairs: Chinese,
Czech, German, Finnish, Russian, Lithuanian and Kazakh to English. For each language, we use the test set that contains several thousands of reference-candidate pairs annotated with human ratings that assess the translation quality.\\

\section{Results}\label{sec:results}

Table \ref{tab:evaluation} reports the results averaged over the 4 TAC datsets and the 7 WMT to-English language pairs.
Details are provided in Appendices \ref{app:a} and \ref{app:b}.

We benchmarked the metrics in terms of Pearson correlations with human scores, runtimes, and numbers of parameters.
We used two approaches to compute the Pearson correlations: summary-level (or segment-level) and system-level.

In the former approach, a score is attributed to each of the output candidates, while in the latter approach, one single overall score is attributed to the system (by averaging its individual scores). 

Rows \texttt{a} to \texttt{c} correspond to BERTScore with BERT miniatures as the underlying model.
They are simple baselines added for the sake of comparison, to see what we get when BERTScore is used with the same number of parameters as FrugalScore.

Rows \texttt{d} to \texttt{g} correspond to the expensive metrics that are learned by FrugalScore (in the respective sections of the bottom half of the table). They are BERTScore and MoverScore metrics where the underlying model is a large pretrained language model: BERT-Base ($L=12$, $H=512$), RoBERTa-Large ($L=24$, $H=1024$) \citep{liu2019roberta}, and DeBERTa-XLarge ($L=24$, $H=1536$) \citep{he2020deberta}.

Finally, rows \texttt{i} to \texttt{xii} correspond to FrugalScore.
Subscripts refer to row labels and indicate which metric-model combination was used to annotate pairs.
I.e., FrugalScore$_d$ learned the metric of row $d$, i.e., BERTScore with BERT-Base.

Note that FrugalScores were only created for the metrics relying on large models (rows \texttt{d} to \texttt{g}).
Rows \texttt{a} to \texttt{c}, as was already explained, are just for sanity checking.

First, results show that all FrugalScores, regardless of which metric they learned, significantly outperform the BERTScores with miniature models.
These results suggest that FrugalScore is a better approach than using an existing metric with an already-compressed underlying model.
The reason for this is probably that in FrugalScore, the knowledge of the original unsupervised metric (based on a large model) is explicitly transferred to the miniature via the continuation of its pretraining on the synthetic dataset. That is, the miniature is actually learning a metric. Whereas, on the other hand, plugging a compressed version of a general-purpose language model into the original unsupervised metric just makes it lose expressiveness and capacity.

Second, we can clearly see that FrugalScore retains most of the performance of the original metric, while running several times faster and reducing the number of parameters by several orders of magnitude.
On average over all metrics, tasks, and miniatures, FrugalScore retains 96.8\% of the original performance, runs 24 times faster, and has 35 times less parameters. 

More precisely, on average across all metrics, FrugalScore-Tiny retains 97.7/94.7\% of the original performance on TAC (pyramid score/responsiveness), while running 54 times faster and having 84 times less parameters.
Its small and medium versions retain near full performance in terms of responsiveness (98 and 97.7\%) and even slightly outperform the original metrics in terms of pyramid score, while at the same time reducing the runtime and the number of parameters by 32 (resp. 21) and 13 (resp. 9) times.

On WMT, FrugalScore-Tiny retains 88.58\% of the performance of the original metrics, while running 14 times faster (and still having 84 times less parameters), while the small and medium versions of FrugalScore retain 95.71 \% and 98.06\% of the original performance while still offering a 32 times (resp. 21) speedup and having 13 times (resp. 9) less parameters, on average.

Interestingly, FrugalScore even improves the performance of the original metrics in some cases.
For example, on TAC, FrugalScore$_g$ with BERT-Tiny (row \texttt{x}) improves the performance of the original MoverScore metric based on BERT-Base (row \texttt{g}) from 66.5 to 67.3 in terms of pyramid score, while reducing the number of parameters by 25 and running 50 times faster.
Other examples, also for TAC with the pyramid score, include FrugalScore$_f$ with BERT-Small (row \texttt{viii}, +1.5 point) and FrugalScore$_f$ with BERT-medium (row \texttt{ix}, +1 point).



Finally, the results of FrugalScore for different miniature sizes show that, on WMT, using larger models always improves performance (e.g., row \texttt{x} $\rightarrow$ \texttt{xi} $\rightarrow$ \texttt{xii}).
But interestingly, on TAC, this observation does not hold (e.g., row \texttt{vi} $\rightarrow$ \texttt{viii} $\rightarrow$ \texttt{ix}), and sometimes, FrugalScore with the smallest miniature (BERT-Tiny) is superior (e.g. rows \texttt{i} and \texttt{x}).
This finding suggests that the impact of the pretrained language model size is task-dependent.

To sum up, results clearly show the effectiveness of FrugalScore in learning a cheaper, lighter, and faster version of the original metrics, while retaining most of their original performance.
The system-level correlations, provided in Appendices \ref{app:c} and \ref{app:d}, corroborate these positive results.


\begin{center}
\begin{table*}[ht]
\small
\def\arraystretch{1.5}
\begin{center}
\begin{tabular}{ |c|c|c|c|c|c|c|c|} 
\hline
& \makecell{Pretraining \\ Continued}  &  TAC-2008 & TAC-2009 & TAC-2010 & TAC-2011 & Average \\
\hline
\multirow{2}{4.5em}{TAC-08} & no &  \multirow{2}{*}{-}  & 67.7$_{0.57}$ & 66.1$_{0.18}$ & 63.6$_{0.36}$ & 65.8\\
  & yes &  & 74.4$_{0.13}$& 71.3$_{0.04}$ & 67.3$_{0.13}$ & 71.0 \\
\hline
\multirow{2}{4.5em}{TAC-09} & no & 61.4$_{0.41}$ & \multirow{2}{*}{-} & 66.9$_{0.24}$ & 62.7$_{0.55}$ & 63.7\\
  & yes & 65.8$_{0.25}$ & & 70.7$_{0.32}$ & 66.0$_{0.18}$ & 67.5 \\
 \hline
\multirow{2}{4.5em}{TAC-10} & no &  59.7$_{0.47}$ & 67.3$_{0.7}$ & \multirow{2}{*}{-} & 62.4$_{0.47}$ & 63.1 \\
  & yes & 64.7$_{0.19}$ & 74.3$_{0.24}$ & & 67.2$_{0.11}$ & 68.7 \\
 \hline
 \multirow{2}{4.5em}{TAC-11} & no &  57.6$_{1.39}$ & 64.7$_{1.03}$ & 66.5$_{0.66}$ &  \multirow{2}{*}{-}  & 62.9 \\
  & yes & 63.9$_{0.31}$ & 72.0$_{0.44}$ & 71.6$_{0.44}$ & & 69.2\\
\hline
\end{tabular}
\end{center}
\caption{\label{tab:supervised} Summary-level Pearson correlations with human judgments (Pyramid scores), averaged over 3 runs (standard deviation as subscript). Rows correspond to the training sets and columns to the test sets.}
\end{table*}
\end{center}

\section{Fine-tuning on Human Annotations}\label{section:supervised}
We test two hypotheses in this section: (1) whether fine-tuning on a human-annotated dataset is beneficial, and (2) when fine-tuning on human annotations, whether continuing pretraining on our synthetic dataset is useful.

Because we cannot use the same dataset for fine-tuning and evaluation, we fine-tune a BERT-Small on each year of TAC 2008-2011 for 4 epochs, using two other years as the validation set, and the remaining year as the test set.
The best epoch is selected based on validation performance.
We use a batch size of 32 and a learning rate of 2e-5 that linearly decreases to zero.
Finally, we experiment with two scenarios: fine-tuning the miniature directly without continuing its pretraining on our synthetic dataset, and fine-tuning it after the pretraining continuation (with annotations generated by BERTScore-BERT-Base).

\noindent \textbf{Results}.
Results are reported in Table \ref{tab:supervised} in terms of summary-level Pearson correlations with human evaluations (Pyramid), averaged over 3 runs with different random seeds.

First, it is obvious that everywhere, continuing the pretraining on our synthetic dataset leads to a significant boost in performance.
This is in accordance with \citet{sellam2020bleurt}, who found that pretraining was beneficial even in a supervised setting.

Second, even if a direct comparison is not possible, we can remark when looking at the TAC Pyramid score of row ii) in Table \ref{tab:evaluation} (FrugalScore$_{d}$-BERT-Small) that fine-tuning after pretraining seems very beneficial too.
Indeed, after fine-tuning, we reach on average 71, 67.5, 68.7, and 69.2 (depending on the split), which represents overall a gain of 4.4 points over the non-fine-tuned model (score of 64.7).

\section{Impact of Data Sources}\label{sec:ablation}
To test the importance of each data source introduced in subsection \ref{sub:synth}, we created a training set containing sequence pairs uniformly and equally sampled from each source.
We annotated these pairs with the BERTScore-BERT-Base metric and we used them to continue the pretraining of a BERT-Small miniature.

We also considered pairs drawn at random from the pairs generated with the other strategies.
The motivation for random pairs was to sample ``negative examples'', as seeing only ``positive examples'' (pairs of related sequences) could bias the learned metric towards considering any two unrelated sequences as similar.

\begin{figure}
\includegraphics[width=7cm]{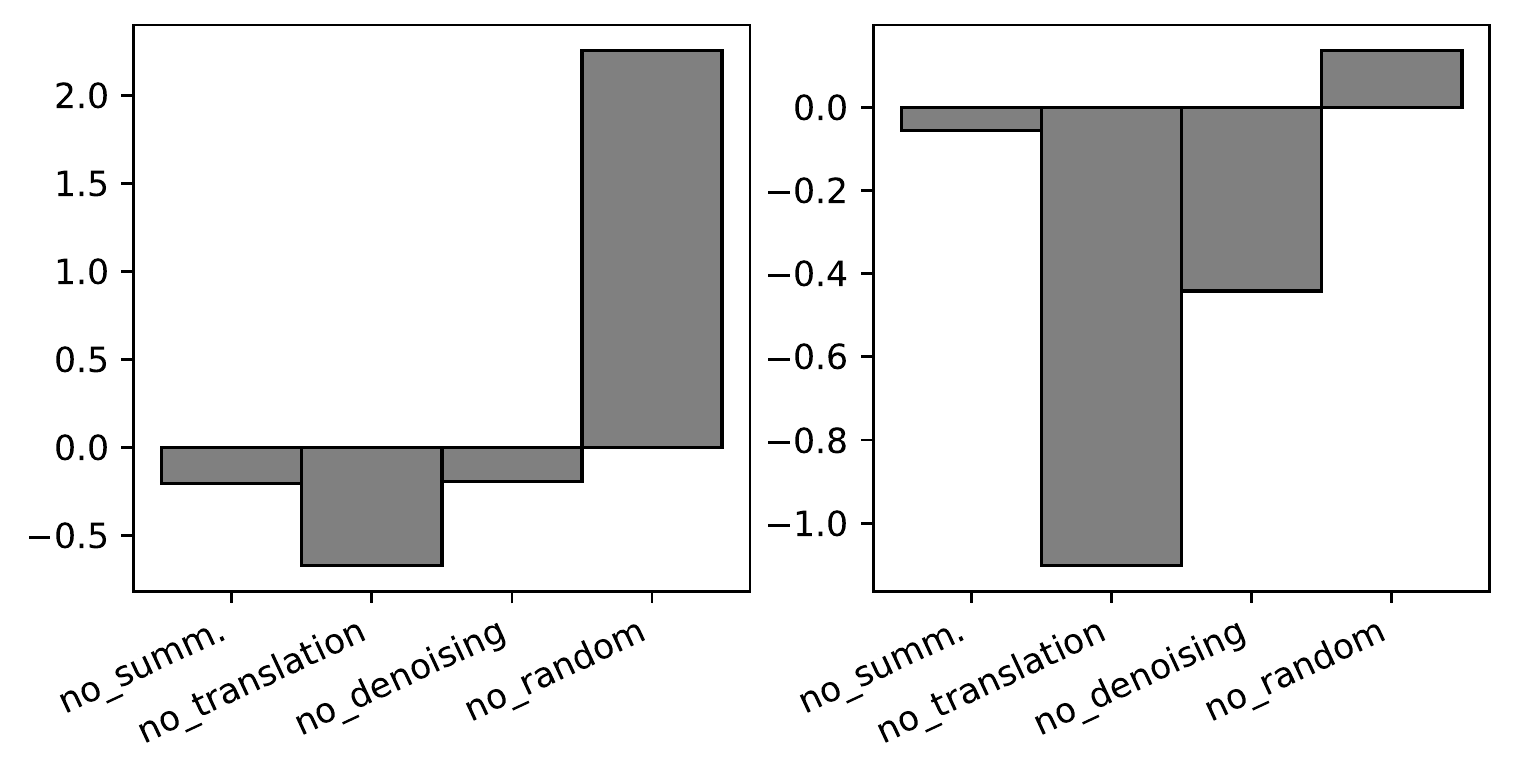}
\caption{Relative improvement in Pearson correlation compared to a dataset covering all sources. Left: TAC. Right: WMT.}
\label{figure:ablation}
\end{figure}

We then continued the pretraining of the BERT-Small miniature four times, excluding each time the pairs coming from a specific data source.
We evaluated the learned metric on TAC-2008 to 2011 and on WMT-2019.
Figure \ref{figure:ablation} shows the average improvements in the Pearson correlation with human judgments relative to training a model on all sources.
Note that when training on all four sources, we sampled 30k pairs from each source (120k total), and when excluding a source, we sampled 40k pairs from each source (120k total).

We can clearly see that excluding the random pairs improves performance while excluding any of the other data sources decreases performance.
In other words, all our data sources are beneficial, and it is not necessary to add ``negative examples''.
We hypothesise that this is due to the fact that NLG datasets typically do not contain completely unrelated pairs of sentences.
Interestingly, the pairs generated with the backtranslation strategy have the greatest impact on performance.
\\

\section{Conclusion}
We proposed FrugalScore, an approach to learn a fixed, low-cost version of any expensive NLG evaluation metric.
Experiments on summarization and translation tasks show that our FrugalScore versions of BERTScore and MoverScore retain most of the original performance in terms of the correlation with human judgments, while running several times faster and having several orders of magnitude less parameters.
On average over all learned metrics, tasks, and variants, FrugalScore retains 96.8\% of the  performance, runs 24 times faster, and has 35 times less parameters than the original metrics.

\bibliography{custom}
\bibliographystyle{acl_natbib}
\clearpage

\appendix
\onecolumn
\section{Detailed TAC Evaluation per Year} \label{app:a}
\begin{table}[ht]
\scriptsize
\def\arraystretch{1.4}
\begin{center}
\begin{tabular}{  |c|c|c|c|c|c|c|c|c|c|c|c|c|} 
\hline
 & Metric & Model & TAC-2008 & TAC-2009 & TAC-2010 & TAC-2011 & \makecell{Macro Avg. \\ Score} & Runtime & Params \\
\hline
a & BERTScore & BERT-Tiny & 52.1/44.4 & 62.2/51.9 & 54.6/49.9 & 52.7/43.6 & 55.4/47.5 & 1m 27s & 4.4M   \\
b & BERTScore & BERT-Small & 56.0/47.8 & 70.0/54.6 & 61.1/54.5 & 59.1/49.2 & 61.6/51.5 & 2m 20s & 29.1M\\
c & BERTScore & BERT-Medium & 57.3/48.5 & 70.6/55.3 & 63.1/56.2 & 59.7/49.5 & 62.7/52.4 & 2m 28s & 41.7M \\ \hline
d & BERTScore & BERT-Base & 61.3/52.2 & 73.2/58.7 & 63.3/56.8 & 61.0/51.2 & 64.7/54.7 & 3m 28s & 110M\\
e & BERTScore & RoBERTa-Large & 56.4/50.9 & 71.1/58.3 & \textbf{69.1}/\textbf{61.4} & 60.3/50.8 & 64.2/55.4 & 5m 17s & 355M \\
f & BERTScore & DeBERTa-XLarge & 60.9/\textbf{54.5} & \textbf{73.9}/\textbf{60.4} & 62.6/56.0 & 61.5/\textbf{53.0} & 64.5/\textbf{56.0} & 6m 20s & 900M \\
g & MoverScore & BERT-Base & \textbf{64.7}/54.2 & \textbf{73.9}/58.2 & 64.7/57.0 & \textbf{62.6}/52.5 & \textbf{66.5}/55.4 & 301m 29s & 110M \\
\hline \hline
i & FrugalScore$_d$ & BERT-Tiny & 60.9/50.0 & 72.5/56.4 & 64.8/57.5 & 61.4/50.0 & 64.9/53.5 & 1m 28s & 4.4M \\
ii & FrugalScore$_d$ & BERT-Small  & 61.9/51.8 & 73.0/57.3 & 62.6/55.8 & 61.3/50.0 & 64.7/53.7 & 1m 35s & 29.1M \\
iii & FrugalScore$_d$ & BERT-Medium & 62.0/52.2 & 73.3/\textbf{58.1} & 62.6/56.0 & 61.3/50.6  & 64.8/54.2 & 1m 55s &  41.7M \\
\hline
iv & FrugalScore$_e$ & BERT-Tiny & 54.8/46.4 & 66.8/54.2 & 61.8/53.1 & 56.4/46.7 & 60.0/50.1 & 1m 28s & 4.4M \\
v & FrugalScore$_e$ & BERT-Small & 59.1/49.6 & 72.7/55.7 & \textbf{68.1}/59.8 & 63.0/50.1 &  64.1/53.8 & 2m 29s & 29.1M \\
vi & FrugalScore$_e$ & BERT-Medium & 57.9/48.4 & 71.8/54.4 & 65.7/57.0 & 60.3/48.5 & 63.9/52.1 & 3m 41s & 41.7M \\
\hline
vii & FrugalScore$_f$ & BERT-Tiny & 57.8/48.5 & 68.6/55.7 & 63.0/54.8 & 57.5/47.8 & 61.7/51.0 & 1m 28s &4.4M \\
viii & FrugalScore$_f$ & BERT-Small & 60.1/51.0 & 73.5/57.5 & 67.3/59.5 &  63.1/51.7 & 66.0/54.9 & 2m 29s & 29.1M\\
ix & FrugalScore$_f$ & BERT-Medium & 59.0/50.3 & 73.3/57.4 & 67.2/\textbf{60.2} & 62.4/51.5 & 65.5/54.9 & 3m 41s & 41.7M \\
\hline
x & FrugalScore$_g$ & BERT-Tiny & 63.6/51.7 & \textbf{74.4}/57.3 & 68.0/60.1 & \textbf{63.2}/51.2 & \textbf{67.3}/\textbf{55.1} & 1m 28s & 4.4M\\
xi & FrugalScore$_g$ & BERT-Small & 63.2/52.5 & 73.1/57.1 & 65.1/57.6 & 62.3/51.5 & 65.9/54.7 & 2m 29s & 29.1M\\
xii & FrugalScore$_g$ & BERT-Medium  & \textbf{63.8}/\textbf{53.2} & 73.6/57.7 & 65.3/57.5 & 62.1/\textbf{51.8} & 66.2/\textbf{55.1} & 3m 41s & 41.7M \\
\hline
\end{tabular}
\end{center}
\caption{Summary-level Pearson correlation (pyramid score/responsiveness).}
\end{table}

\section{Detailed WMT Evaluation per Language} \label{app:b}
\begin{table}[ht]
\scriptsize
\def\arraystretch{1.4}
\begin{center}
\begin{tabular}{  |c|c|c|c|c|c|c|c|c|c|c|c|c|} 
\hline
  & Metric & Model & de-en &  fi-en & gu-en & kk-en & lt-en & ru-en & zh-en &  \makecell{Macro Avg. \\ Score} & Runtime & Params \\
\hline
a & BERTScore & BERT-Tiny & 29.7 & 32.5 & 33.9 & 52.0 & 40.5 & 30.7 & 44.2  & 37.6 & 1m 22s & 4.4M\\
b & BERTScore & BERT-Small & 30.0 & 33.6 & 34.6 & 52.4 & 42.3 & 31.8 & 49.1 & 39.1 & 1m 42s & 29.1M \\
c & BERTScore & BERT-Medium & 30.8 & 34.4 & 35.2 & 52.8 & 42.8 & 32.4 & 50.3 & 39.8 & 2m 04s & 41.7M \\ \hline
d & BERTScore & BERT-Base & 32.8 & 37.4 & 37.1 & 54.0 & 44.7 & 33.7 & 53.7 & 41.9 & 2m 09s & 110M  \\
e & BERTScore & RoBERTa-Large  & 35.3 & 38.7 & 38.7 & 52.0 &  45.3 & 34.3 & \textbf{58.3} & 43.2 & 3m 03s & 355M \\
f & BERTScore & DeBERTa-XLarge & \textbf{37.6} & \textbf{39.2} & \textbf{40.3} & 53.4 & \textbf{47.3} & \textbf{35.7} & 57.8 & \textbf{44.5} & 3m 49s & 900M \\
g & MoverScore & BERT-Base & 36.5 & 39.1 & 39.3 & \textbf{55.0} & 46.5 & 35.6 & 56.0 & 44.0 & 64m 32s & 110M \\
\hline\hline
i & FrugalScore$_d$ & BERT-Tiny & 30.2 & 32.8 & 34.6 & 52.4 & 39.9 & 31.2 & 47.7 & 38.4 & 1m 18s & 4.4M \\
ii & FrugalScore$_d$ & BERT-Small & 32.6 & 35.9 & 37.1 & 54.1 & 43.5 & 33.6 & 52.3 & 41.3 & 1m 35s & 29.1M\\
iii & FrugalScore$_d$ & BERT-Medium & 32.9 & 37.0 & 37.4 & 54.4 & 44.3 & 34.1 & 53.2 & 41.9 & 1m 55s & 41.7M \\
\hline
iv & FrugalScore$_e$ & BERT-Tiny & 30.6 & 32.8 & 33.0 & 49.8 & 38.7 & 29.8 & 48.1 & 37.5 & 1m 18s & 4.4M \\
v & FrugalScore$_e$ & BERT-Small & 33.7 & 35.4 & 35.4 & 51.6 & 42.6 & 32.6 & 52.5 & 40.5 & 1m 35s & 29.1M  \\
vi & FrugalScore$_e$ & BERT-Medium & 35.2 & 37.1 & 35.6 & 52.0 & 44.0 & 33.8 & 54.4 & 41.7 & 1m 55s & 41.7M \\
\hline
vii & FrugalScore$_f$ & BERT-Tiny &  30.8 & 33.1 & 34.4 & 50.8 & 39.4 & 30.4 & 47.1 & 38.0 & 1m 18s & 4.4M \\
viii & FrugalScore$_f$ & BERT-Small & 34.5 & 36.4 & 37.0 & 52.7 & 43.9 & 33.4 & 52.6 & 41.5 & 1m 35s & 29.1M \\
ix & FrugalScore$_f$ & BERT-Medium & 35.8 & \textbf{38.3} & 37.7 & 53.4 & 45.7 & 34.8 & \textbf{55.1} & 43.0 & 1m 55s & 41.7M \\
\hline
x & FrugalScore$_g$ & BERT-Tiny & 33.0 & 34.0 & 36.2 & 53.6 & 40.5 & 32.7 & 48.6 & 39.8 & 1m 18s & 4.4M \\
xi & FrugalScore$_g$ & BERT-Small & 35.6 & 37.4 & 38.9 & 55.0 & 44.8 & 34.8 & 52.8 & 42.8 & 1m 35s & 29.1M \\
xii & FrugalScore$_g$ & BERT-Medium & \textbf{36.2} & \textbf{38.3} & \textbf{39.1} & \textbf{55.6} & \textbf{45.8} & \textbf{35.3} & 54.7 & \textbf{43.6} & 1m 55s & 41.7M \\
\hline

\end{tabular}
\end{center}
\caption{Segment-level Pearson correlation.}
\end{table}
\clearpage
\section{Detailed TAC Evaluation per Year (System Level)} \label{app:c}
\begin{table}[ht]
\scriptsize
\def\arraystretch{1.4}
\begin{center}
\begin{tabular}{  |c|c|c|c|c|c|c|c|c|c|} 
\hline
 & Metric & Model & TAC-2008 & TAC-2009 & TAC-2010 & TAC-2011  &  \makecell{Macro Avg. \\ Score} & Runtime & Params \\
\hline
a & BERTScore & BERT-Tiny & 82.5/77.6 & 87.4/81.8 & 77.5/75.0 & 82.1/79.2 & 82.4/78.4 & 1m 27s & 4.4M \\
b & BERTScore & BERT-Small & 84.4/81.4 & 95.8/84.0 & 81.3/78.0 & 87.6/85.3 & 87.3/82.2 & 2m 20s & 29.1M \\
c & BERTScore & BERT-Medium & 86.3/82.7 & 96.0/84.6 & 84.0/80.6 & 87.8/85.5 & 88.5/83.3 & 2m 28s & 41.7M \\ \hline
d & BERTScore & BERT-Base & 90.6/87.5 & 96.5/87.5 & 83.7/80.9 & 88.3/86.4 & 89.8/85.6 & 3m 28s & 110M  \\
e & BERTScore & RoBERTa-Large & 80.0/80.9 & 94.7/87.7 & \textbf{92.7}/\textbf{89.8} & 88.9/89.2 & 89.1/86.9 & 5m 17s & 355M \\
f & BERTScore & DeBERTa-XLarge & 88.0/\textbf{89.8} & \textbf{97.5}/\textbf{89.8} & 85.7/84.0 & \textbf{90.7}/\textbf{91.8} & 90.5/\textbf{88.9} & 6m 20s & 900M \\
g & MoverScore & BERT-Base & \textbf{95.4}/89.5 & 96.9/85.9 & 85.7/84.0 & 88.6/86.0  & \textbf{91.7}/86.3 & 301m 29s & 110M \\
\hline \hline
i & FrugalScore$_d$ & BERT-Tiny & 91.6/85.3 & 95.8/84.7 & 86.2/82.9 & 88.3/84.4  & 90.5/84.3 & 1m 28s & 4.4M \\
ii & FrugalScore$_d$ & BERT-Small  & 90.9/86.8 & 96.2/85.4 & 82.8/79.6 & 87.8/84.3 & 89.4/84.0 & 1m 35s & 29.1M \\
iii & FrugalScore$_d$ & BERT-Medium & 90.6/87.0 & 96.6/86.3 & 82.5/79.6 & 87.6/84.9 & 89.3/84.5 & 1m 55s &  41.7M \\
\hline
iv & FrugalScore$_e$ & BERT-Tiny & 86.3/81.1 & 95.1/87.1 & 84.5/80.2 & 84.5/80.9 & 87.6/82.3 & 1m 28s & 4.4M \\
v & FrugalScore$_e$ & BERT-Small & 85.1/81.7 & 95.7/83.6 & \textbf{91.2}/87.5 & 91.7/87.5 & 90.9/85.1 & 2m 29s & 29.1M \\
vi & FrugalScore$_e$ & BERT-Medium & 81.6/80.7 & 95.7/84.1 & 90.9/87.5 & 87.6/85.3 & 89.0/84.4 & 3m 41s & 41.7M \\
\hline
vii & FrugalScore$_f$ & BERT-Tiny & 89.7/84.5 & 95.3/\textbf{87.6} & 85.1/81.4 & 84.8/81.2 & 88.7/83.7 & 1m 28s &4.4M \\
viii & FrugalScore$_f$ & BERT-Small & 86.8/85.1 & 96.7/85.4 & 89.5/86.2 &  91.6/88.7 & 91.2/86.3 & 2m 29s & 29.1M \\
ix & FrugalScore$_f$ & BERT-Medium & 85.4/86.3 & \textbf{97.2}/87.2 & 91.1/\textbf{88.9} & \textbf{92.3}/\textbf{91.0} & 91.5/\textbf{88.3} & 3m 41s & 41.7M \\
\hline
x & FrugalScore$_g$ & BERT-Tiny & \textbf{93.7}/86.1 & 96.2/83.9 & 90.1/87 & 89.4/84.8 & \textbf{92.3}/85.5 & 1m 28s & 4.4M \\
xi & FrugalScore$_g$ & BERT-Small & 93.2/\textbf{87.6} & 96.4/84.2 & 85/81.7 & 87.9/84.9 & 90.6/84.6 & 2m 29s & 29.1M \\
xii & FrugalScore$_g$ & BERT-Medium  & \textbf{93.7}/87.5 & 96.5/84.5 & 84.8/81.6 & 87.3/84.7 & 90.6/84.6 & 3m 41s & 41.7M \\
\hline
\end{tabular}
\end{center}
\caption{System-level Pearson correlation (pyramid/responsiveness).}
\end{table}

\section{Detailed WMT Evaluation per Language (System Level)} \label{app:d}
\begin{table}[ht]
\scriptsize
\def\arraystretch{1.4}
\begin{center}
\begin{tabular}{  |c|c|c|c|c|c|c|c|c|c|c|c|c|} 
\hline
  & Metric & Model & de-en &  fi-en & gu-en & kk-en & lt-en & ru-en & zh-en  & \makecell{Macro Avg. \\ Score} & Runtime & Params\\
\hline
a & BERTScore & BERT-Tiny & 74.1 & 97.9 & 93.1 & 99.77 & 87.9 & 94.5 & 91.7 & 91.3 & 1m 22s & 4.4M \\
b & BERTScore & BERT-Small & 82.6 & 97.5 & 88.2 & \textbf{99.87} & 95.3 & 96.4 & 93.0 & 93.3 & 1m 42s & 29.1M \\
c & BERTScore & BERT-Medium & 83.7 & 97.7 & 88.2 & 99.86 & 94.4 & 96.2 & 93.5 & 93.4 & 2m 04s & 41.7M  \\ \hline
d & BERTScore & BERT-Base & 89.1 & 97.8 & 89.7 & 99.72 & 96.9 & 96.9 & 95.8  & 95.1 & 2m 09s & 110M \\
e & BERTScore & RoBERTa-Large  & \textbf{94.0} & 98.4 & 98.1 & 98.00 &  96.1 & 91.0 & 98.2 & 96.3 & 3m 03s & 355M \\
f & BERTScore & DeBERTa-XLarge & 93.9 & 98.3 & \textbf{98.2} & 99.18 & \textbf{98.7} & 97.1 & \textbf{98.4} & \textbf{97.7} & 3m 49s & 900M \\
g & MoverScore & BERT-Base & 88.1 & \textbf{99.1} & 91.2 & 98.58 & 96.0 & \textbf{97.2} & 96.4 & 95.2 & 64m 32s & 110M  \\
\hline\hline
i & FrugalScore$_d$ & BERT-Tiny & 81.1 & 98.6 & 94.4 & 99.80 & 92.2 & 95.4 & 93.8 & 93.6 & 1m 18s & 4.4M \\
ii & FrugalScore$_d$ & BERT-Small & 86.5 & 98.5 & 93.6 & 99.82 & 95.9 & 97.1 & 94.7 & 95.2 & 1m 35s & 29.1M \\
iii & FrugalScore$_d$ & BERT-Medium & 88.3 & 98.3 & 92.1 & 99.79 & 96.4 & 97.2 & 95.4 & 95.4 & 1m 55s & 41.7M \\
\hline
iv & FrugalScore$_e$ & BERT-Tiny & 80.2& 97.7 & 94.9 & 99.73 & 86.4 & 94.6 & 93.7 & 92.5 & 1m 18s & 4.4M \\
v & FrugalScore$_e$ & BERT-Small & 83.9 & 98.0 & 95.2 & 99.79 & 92.4 & 97.0 & 95.1 & 94.5 & 1m 35s & 29.1M \\
vi & FrugalScore$_e$ & BERT-Medium & 88.1 & 97.9 & 93.0 & 99.78 & 94.9 & \textbf{97.8} & 96.1 & 95.4 & 1m 55s & 41.7M \\
\hline
vii & FrugalScore$_f$ & BERT-Tiny &  81.3 & 97.9 & 96.1 & 99.81 & 89.8 & 94.7 & 93.7 & 93.3 & 1m 18s & 4.4M \\
viii & FrugalScore$_f$ & BERT-Small & 85.8 & 97.7 & \textbf{96.2} & \textbf{99.85} & 95.3 & 97.3 & 95.7 & 95.4 & 1m 35s & 29.1M \\
ix & FrugalScore$_f$ & BERT-Medium & \textbf{89.9} & 97.9 & 90.8 & \textbf{99.85} & \textbf{97.6} & \textbf{97.8} & \textbf{96.9} & \textbf{95.8} & 1m 55s & 41.7M \\
\hline
x & FrugalScore$_g$ & BERT-Tiny & 81.8 & \textbf{98.9} & 95.6 & 99.73 & 92.1 & 95.6 & 94.4 & 94.0 & 1m 18s & 4.4M \\
xi & FrugalScore$_g$ & BERT-Small & 85.4 & 98.8 & 95.8 & 99.52 & 94.9 & 96.8 & 95.3 & 95.2 & 1m 35s & 29.1M \\
xii & FrugalScore$_g$ & BERT-Medium & 87.0 & 98.8 & 93.5 & 99.29 & 95.6 & 97.0 & 95.9 & 95.3 & 1m 55s & 41.7M \\
\hline

\end{tabular}
\end{center}
\caption{System-level Pearson correlation.}
\end{table}

\end{document}